# Improved Bitcoin Price Prediction based on COVID-19 data

**Palina Niamkova[1*], Rafael Moreira[2]**
[1,2]University of North Texas, USA
poli.nemkova@unt.edu,  RafaelAlvesMoreira@my.unt.edu

## Abstract

Social turbulence can affect people's financial decisions, causing changes in spending and saving. During a global turbulence as significant as the COVID-19 pandemic, such changes are inevitable.
Here we examine how the effects of COVID-19 on various jurisdictions influenced the global price of Bitcoin. We hypothesize that lockdowns and expectations of economic recession erode people's trust in fiat (government-issued) currencies, thus elevating cryptocurrencies. Hence, we expect to identify a causal relation between the turbulence caused by the pandemic, demand for Bitcoin, and ultimately its price.
To test the hypothesis, we merged datasets of Bitcoin prices and COVID-19 cases and deaths. We also engineered extra features and applied statistical and machine learning (ML) models. We applied a Random Forest model (RF) to identify and rank the feature importance, and ran a Long Short-Term Memory (LSTM) model on Bitcoin prices dataset twice: with and without accounting for COVID-19 related features.
We find that adding COVID-19 data into the LSTM model improved prediction of Bitcoin prices.

## 1  Introduction

The COVID-19 pandemic influences many aspects of today's life. Growing numbers of new cases and deaths, unprecedented lockdown, and irrational leaders' responses synthesize an overwhelming uncertainty.  As a result, we can see a nontrivial response from the current population. For example, since the beginning of the pandemic the sales of guns and ammo have increased dramatically [Acorn, 2020]. Hence, it suggests that people are opting in for the less conventional means to get feelings of safety in the pandemic's uncertain times.
We suggest that similar behavior may be found in other areas, for example in the search for money safety. As a result of the social turmoil people may lose trust in sate-controlled assets and move to a more novel and independent way to hold their investment – cryptocurrencies. Additionally, possible change in the use of cryptocurrencies may be caused by other, more malicious, motivations that are still related to the pandemic. For illegal markets the freeze of international travel may lead to a use alternative ways of money transfer.
In this paper we are exploring our hypothesis that COVID-19 pandemic influences the prices of cryptocurrencies (Bitcoin). We are applying machine learning models to forecast the prices of Bitcoin with and without accounting for COVID-19.

## 2  Related Work

Researchers in many fields are interested to investigate the possible effect of such a nontrivial social turmoil as the global COVID-19 pandemic has caused. It is not surprising to see the growing number of research papers of this nature in finance implications. Relationships between the pandemic and prices of assets is a popular discussion topic among financial scholars and researchers.
One discussion is focused on the change in returns on the assets such as cryptocurrencies. Conlon and McGee in their 2020 paper conclude that Bitcoin does not pass a heaven safety test [Conlon, 2020] or exhibit high volatility.
Other discussions focus on the relationships between COVID-19 and the assets.
For example, Demir et al. [Demir, 2020] investigates the relationship between COVID-19 and cryptocurrencies. The researchers found that Bitcoin initially had a negative correlation but on a later stage the correlation turned to be positive. Going further, in another paper Goodell & Goutte determine that COVID-19 caused a rise in Bitcoin prices [Goodell, 2020].
At this point in time the available COVID-19 pandemic related research in finance and machine learning is still rare but highly relevant. Therefore, we are aiming to add to the academic discussion with our contribution by examing the effect of the COVID-19 pandemic on the cryptocurrency market.

## 3  Data acquiring and preprocessing

We acquired our Bitcoin dataset using the Bitfinex API [Klein, 2019]. The data consisted of values per minute for the following features: open price, close price, and volume. Our data for the COVID-19 pandemic was acquired from the World Health Organization [WHO, 2020]. This data contained the following daily measures: cumulative number of cases, number of new cases per 24 hours, cumulative number of deaths, and number of new deaths per last 24 hours. Each of the datasets had corresponding values starting from January 6, 2020 and until September 5, 2020. We also engineered additional features by calculating mean, kurtosis, and skew for each feature. This step also helped to address the issue of different number of samples in the Bitcoin dataset (which had samples per minute) and the COVID-19 dataset with daily values. The final merged dataset had values per day and total of 246 samples of data with 37 features. Our dataset and code are available for public on Github [Niamkova, 2020].

## 4 Model Setup

We performed our calculations using Python in a Google Colab notebook and stored our version-controlled code and data on Github.

First, we intended to identify the most sensitive features out of the 37 that we were gathered. Since the initial dataset for COVID-19 was relatively small (less than a year of data), we wanted to reduce number of features in order to prevent overfitting the model. We started with applying a Random Forest (RF) model to get the ranked feature importance list.

We used the scikit-learn library to get the Random Forest Regressor function. Our RF model with all features had a mean absolute error of 37.17 and an accuracy of 99.49 %. After running this model using important features only, we got a mean absolute error of 22.6 and an accuracy of 99.7 %. The RF model identified seven important features that are related to the Bitcoin market. We continued with the planned statistical approach.

To identify more features, we ran a Pearson correlation and analyzed the correlation matrix together with p-values of the features. After identifying features with a correlation higher than 0.6 and a p-value lower than 0.05 we added 12 more features. The final pre-procced dataset contains 19 features.

After exploring the data with statistical tools, we continued with the trainable model: the Long Short-Term Memory model (LSTM). We opted for a LSTM model above other deep learning models due to its superior performance in handling sequence dependence such as time series [Livieris, 2020]. We employed the LSTM model through Keras – a neural network Python library. We split the data for training, testing, and validation in the following ratio respectively: 70%-20%-10%.

## 5 Results

In our model we use windows of data to make sets of predictions for 1 day based on 23 days of data. Hence, our prediction horizon is one day, and window width is 24 days.

After training the LSTM model on the dataset with Bitcoin prices only there was a mean absolute error of 0.9034 with a loss of 1.4127.

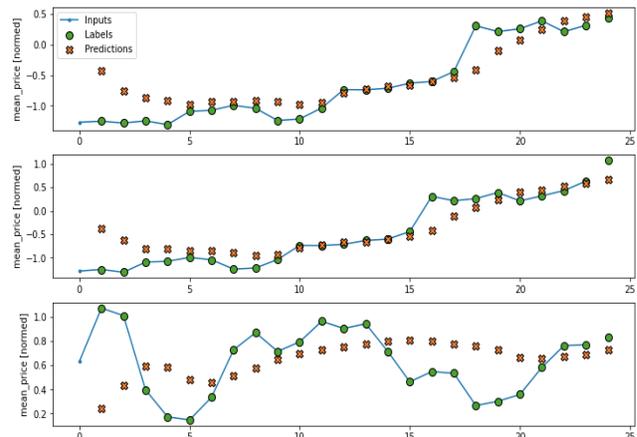

Fig 1. LSTM model set of predictions of daily Bitcoin prices (without COVID-19 features).

Next we trained an LSTM model of the same hyperparameters and architecture but that included the COVID-19 features. After that there was a mean absolute error of 0.7102 with a loss of 0.8492. This suggests that including extra COVID-19 cases may help to predict Bitcoin prices with greater precision.

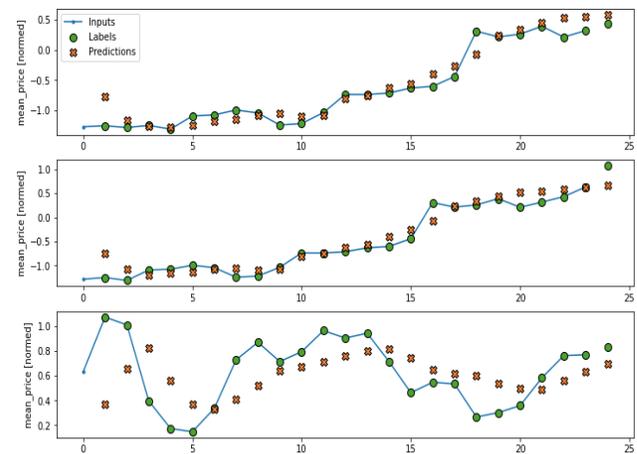

Fig 2. LSTM model prediction of Bitcoin prices including COVID-19 features.

## 6 Discussion

Our current results suggest that COVID-19 has influenced on the market of Bitcoin. Since Bitcoin price is responsive to external shocks [Panagiotodos, 2019], our findings confirm that COVID-19 is a significant external shock.

Our results are also consistent with the paper on behavioral finance where fear sentiment analysis shows the impact of investor sentiment on asset markets [Chen, 2020]. Due to the COVID-19 pandemic's sharply increasing death rate and unprecedented lockdowns and travel bans introducing high negative sentiment [Chen, 2020], the pandemic has impacted Bitcoin prices as well.

Our contribution can help financial managers to improve the robustness of their prediction model for financial portfolio by including COVID-19 features in their deep learning model. Likewise, the academic professionals can use our findings to research more detailed dependencies with the accumulation of more data about COVID-19.

## 7 Conclusion

COVID-19 pandemic has an extraordinary effect on many aspects of our society. We had to change our life and work routines as well as change traveling habits. Our financial preferences are not the least aspect to change is our financial preferences. In this paper we identified that the COVID-19 pandemic has a direct effect on the cryptocurrency market, in particular on Bitcoin prices.